\documentclass[letterpaper, 10 pt, conference]{ieeeconf}  

\IEEEoverridecommandlockouts                              

\usepackage{graphicx} 
\usepackage{amsmath,amssymb,amsfonts}
\usepackage{hyperref}

\usepackage{subcaption}
\newcommand{\squeezeup}{\vspace{-2.5mm}}

\usepackage{color}
\newcommand*\Update{\color{black}}
\newcommand*\UpdateNew{\color{black}}
\newcommand*\Done{\color{black}}

\usepackage{array}
\newcolumntype{C}[1]{>{\centering\let\newline\\\arraybackslash}m{#1}}

\usepackage{booktabs}
\title{\LARGE \bf
Multi Actor-Critic DDPG for Robot Action Space Decomposition: \\ A Framework to Control Large 3D Deformation of Soft Linear Objects
}

\author{Mélodie Daniel$^{1}$, Aly Magassouba$^{2}$, Miguel Aranda$^{3}$, Laurent Lequièvre$^{2}$, Juan Antonio Corrales Ramón$^{4}$, \\ Roberto Iglesias Rodriguez$^{4}$ and Youcef Mezouar$^{2}$
\thanks{$^{1}$Univ. Bordeaux, CNRS, Bordeaux INP, LaBRI, UMR 5800, F-33400 Talence, France. $^{2}$CNRS, Clermont Auvergne INP, Institut Pascal,  Université Clermont Auvergne, Clermont-Ferrand, France. $^{3}$Instituto de Investigación en Ingeniería de Aragón (I3A), Universidad de Zaragoza, Zaragoza, Spain. $^{4}$Centro Singular de Investigación en Tecnoloxías Intelixentes (CiTIUS),  Universidade de Santiago de Compostela, Santiago de Compostela, Spain. JACR was funded by the Spanish government through a ’Beatriz Galindo’ fellowship (Ref. BG20/00143), by the research project PID2020-119367RB-I00 and by the Galician Government through the programme 'Captación e Retención de Talento'. MA was supported via projects PID2021-124137OB-I00 and TED2021-130224B-I00 funded by MCIN/AEI/10.13039/501100011033, by ERDF A way of making Europe and by the European Union NextGenerationEU/PRTR. Corresponding author: Mélodie Daniel, e-mail: \texttt{melodie.daniel@u-bordeaux.fr.}}}

\begin{document}

\maketitle
\thispagestyle{empty}
\pagestyle{empty}

\begin{abstract}
Robotic manipulation of deformable linear objects (DLOs) has great potential for applications in diverse fields such as agriculture or industry. However, a major challenge lies in acquiring accurate deformation models that describe the relationship between robot motion and DLO deformations. Such models are difficult to calculate analytically and vary among DLOs. Consequently, manipulating DLOs poses significant challenges, particularly in achieving large deformations that require highly accurate global models. To address these challenges, this paper presents MultiAC6: a new multi Actor-Critic framework for robot action space decomposition to control large 3D deformations of DLOs. In our approach, two deep reinforcement learning (DRL) agents orient and position a robot gripper to deform a DLO into the desired shape. Unlike previous DRL-based studies, MultiAC6 is able to solve the sim-to-real gap, achieving large 3D deformations up to 40 cm in real-world settings. Experimental results also show that MultiAC6 has a 66\% higher success rate than a single-agent approach. Further experimental studies demonstrate that MultiAC6 generalizes well, without retraining, to DLOs with different lengths or materials. We released the code at this URL\footnote{\url{https://github.com/MelodieDANIEL/MultiAC6}}. A demonstration video is available at this URL\footnote{\url{https://youtu.be/CWyCozJEiQk}}.
\end{abstract}
\section{Introduction} \label{introduction}
Following the Industry 4.0 paradigm, industrial robots are increasingly being requested to manipulate various objects in real-world settings. In this context, providing robots with the ability to manipulate soft objects has many practical uses. This particularly concerns deformable linear objects (DLOs), which are one-dimensional soft objects such as cables, plants, or beams \cite{Sanchez2018}. Typical applications are related to cable harnessing \cite{Zhu2018, Lagneau2020}, hose manipulation \cite{Mitrano2021}, or plant stem bending for harvesting \cite{Botterill2017, Aghajanzadeh2022}. 

Modeling DLOs for robot manipulation remains a challenge. In fact, such objects exhibit nonlinear deformations that are difficult and computationally expensive to accurately model \cite{Yu2023}. Therefore, simplified models are generally used \cite{Zhu2022}, but at the expense of accuracy and flexibility. Indeed, a single deformation model cannot fully capture the length or material of various DLOs.

Different lines of research have tackled DLOs manipulation. Analytical approaches generally consider 2D deformations \cite{Zhu2018, Aghajanzadeh2022, Aghajanzadeh2022Iros, Lv2022, Jin2019}. Fewer works have addressed the more challenging case of 3D deformations \cite{Lagneau2020, Yu2023, Navarro2016, Reza2022}. In general, these methods are limited by the accuracy of the deformation model used. To avoid modeling DLO deformations, another line of research explored deep reinforcement learning (DRL) approaches. Although promising results could be obtained, these approaches are validated mainly in simulation \cite{Laezza2021, Daniel2022, Pecyna2022}. Indeed, the sim-to-real gap, peculiar to DRL approaches, is still an obstacle to real-world applications \cite{Yu2023}. This sim-to-real gap is mainly caused by the approximations of the simulators, such as unrealistic deformations. Despite this limitation, few DRL approaches have been validated in real-world settings, but only for 2D deformations \cite{Han2017, Wu2020}.

\begin{figure}[tb]
\centering
\includegraphics[width=7cm]{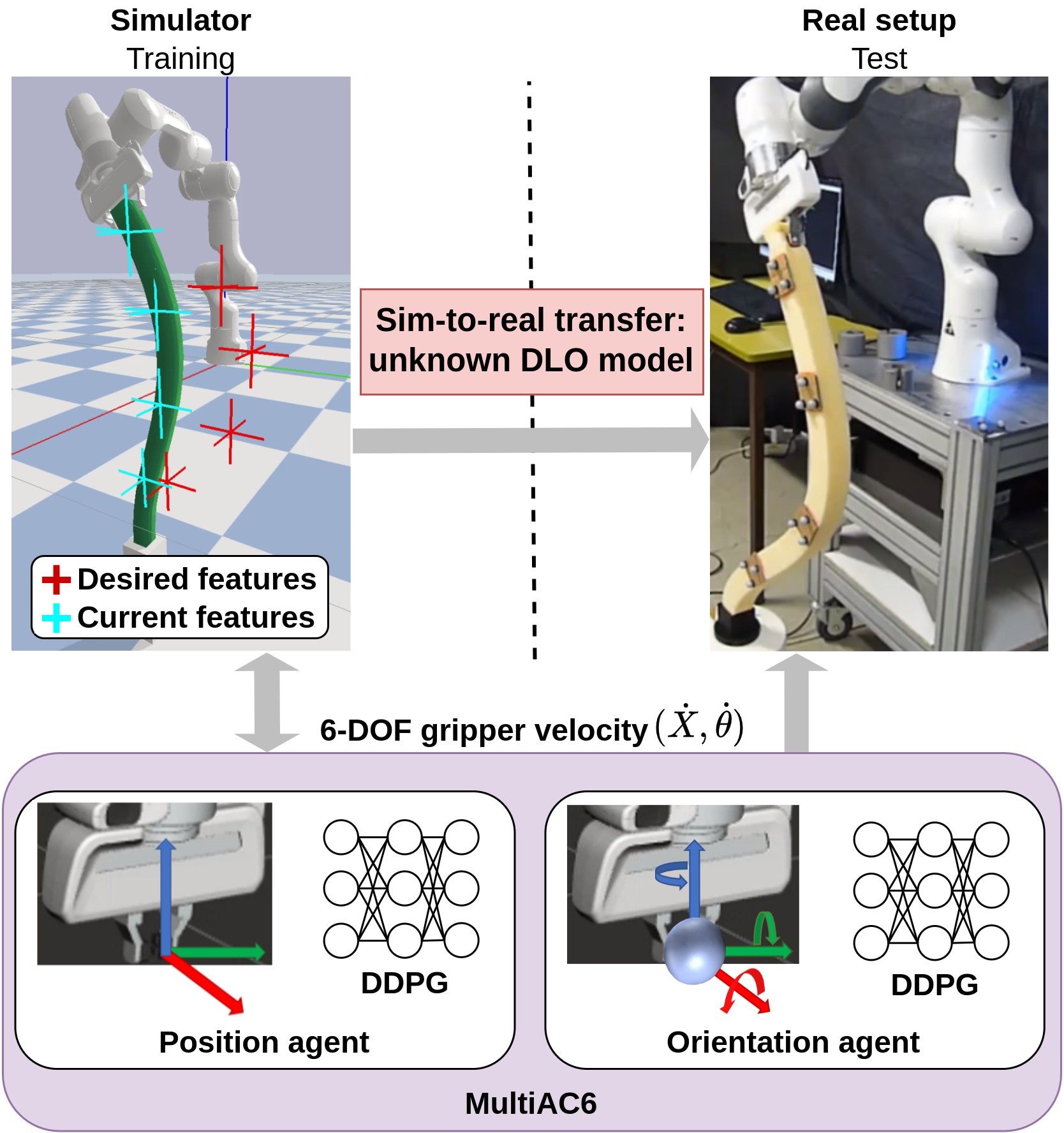}\vspace{-2pt}
\caption{\small Overview of MultiAC6: a multi actor-critic framework  controlling the gripper pose to achieve large 3D DLO deformations.}
\vspace{-0.7cm}
\label{ac6}
\end{figure}

In contrast, this paper addresses the 3D manipulation of DLOs in real-world settings with a single-arm robot. To this end, we propose a novel Multi Actor-Critic (MultiAC6) DRL framework based on the deep deterministic policy gradient (DDPG) algorithm. This framework decomposes the 6 degree of freedom (DOF) action space of the robot gripper to multiple agents, as shown in Figure~\ref{ac6}. This paper is an extension of our previous work \cite{Daniel2022}. In the latter case, we proposed a single agent framework that controls the 3 DOF position of the robot gripper to deform a DLO in simulation. Differently, in this paper, we propose a collaborative multi-agent framework with action space decomposition. Recent work in natural language processing \cite{Wang2021} demonstrated that decomposing action spaces between agents achieves better results than a single-agent framework. Indeed, such an approach reduces the state-action space size significantly and makes exploration in the agent training phase more efficient.

The following key points of MultiAC6 are worth highlighting. First, unlike existing DRL-based approaches \cite{Daniel2022}, MultiAC6 controls the gripper pose (6 DOF) instead of the gripper position (3 DOF). Therefore, the robot can achieve more complex deformations. Second, MultiAC6 overcomes the sim-to-real gap and is experimentally validated. Third, MultiAC6 is robust to DLO variations without retraining or online fine-tuning.
The different contributions of this article can be summarized as follows:
\begin{itemize}
    \item We propose a new DRL collaborative multi actor-critic framework with action space decomposition to address 3D manipulation of DLOs in real-world settings. Our approach consists of two agents controlling the gripper position (3 DOF) and orientation (3 DOF).
    \item We define an optimized reward function based on the maximum error between the current shape of the DLO and its desired shape. This reward performs better than a reward function based on the average error \cite{Daniel2022}.
    \item We validate the robustness of MultiAC6 to DLO variations through extensive real-world experiments involving large 3D deformations. These experiments are carried out, using the same MultiAC6 model, for DLOs with varying length, material, and stiffness.
\end{itemize}

\vspace{-0.2cm}
\section{Related Work}

\Update \textbf{Non-DRL methods}: \Done Several recent works showed 3D shape control of DLOs with dual-arm manipulation \cite{ Lagneau2020, Yu2023, Navarro2016, Reza2022}. The approach in \cite{Reza2022} uses a geometrical model of the object to compute an online Jacobian that guides the control task. In \cite{Lagneau2020, Navarro2016}, quasi-static adaptive controllers based on computing a Jacobian based on a sensor-based deformation model are proposed. To address large 3D deformations, the authors of \cite{Yu2023} combine offline and online learning of a radial basis function network. These studies require an online adaptation for each new DLO used. On the contrary, we achieve generalization to various real-world DLOs without needing online estimations or specific training. Additionally, our setup consists of a single arm, which is more challenging due to the fewer actuated DOFs.

\Update \textbf{DRL methods}: \Done Another branch of research explored DRL-based methods to avoid modeling DLO deformation. These methods are mainly validated in simulation. For example, in \cite{Laezza2021}, a method based on the DDPG algorithm is introduced to control elastoplastic DLOs. In our previous work in \cite{Daniel2022}, we addressed 3D deformations with a DDPG-based architecture. However, such techniques do not offer a way to transfer the learned policies to real-world settings \cite{Yu2023}. 

\Update \textbf{Sim-to-real gap}: \Done Resolving this sim-to-real gap is still an open problem since there are no accurate and standard simulation environments for deformable objects. Most of the existing approaches develop their own environment using a physics-based simulation engine such as Bullet \cite{Seita2021} or Mujoco \cite{Chen2022}. These engines generally model DLOs with the mass-spring method or the finite element method (FEM). This is the case for \cite{Han2017} and \cite{Wu2020} where different DRL agents are trained in customized Mujoco environments. Given the above limitations, these works are among the very few validated in real-world settings. In \cite{Han2017}, a sample-efficient reinforcement learning method named PILCO is proposed to close the sim-to-real gap for 2D deformations. In \cite{Wu2020}, a Soft-Actor-Critic (SAC) algorithm is presented to control 2D deformations of real objects. These contributions are nonetheless limited to 2D deformations for DLOs with no compression strength (i.e., cables, strings, and ropes) \cite{Sanchez2018}. 


Instead, our contribution MultiAC6 aims to close the sim-to-real gap for 3D manipulation of large-strain DLOs, such as elastic tubes. The MultiAC6 action space decomposition is inspired by the multi-agent dialog policy framework proposed in \cite{Wang2021}. Within this framework, each component of the action is carried out by a different agent. In \cite{Wang2021}, this dialog framework was shown to be 11\% more accurate than a hierarchical DRL framework and 66\% more accurate than a single agent framework. 



\section{Problem Statement} \label{PB}

\begin{figure}[tb!]
     \centering
     \hspace{2.5mm}
     \begin{subfigure}[b]{0.2\textwidth}
         \includegraphics[height=5cm]{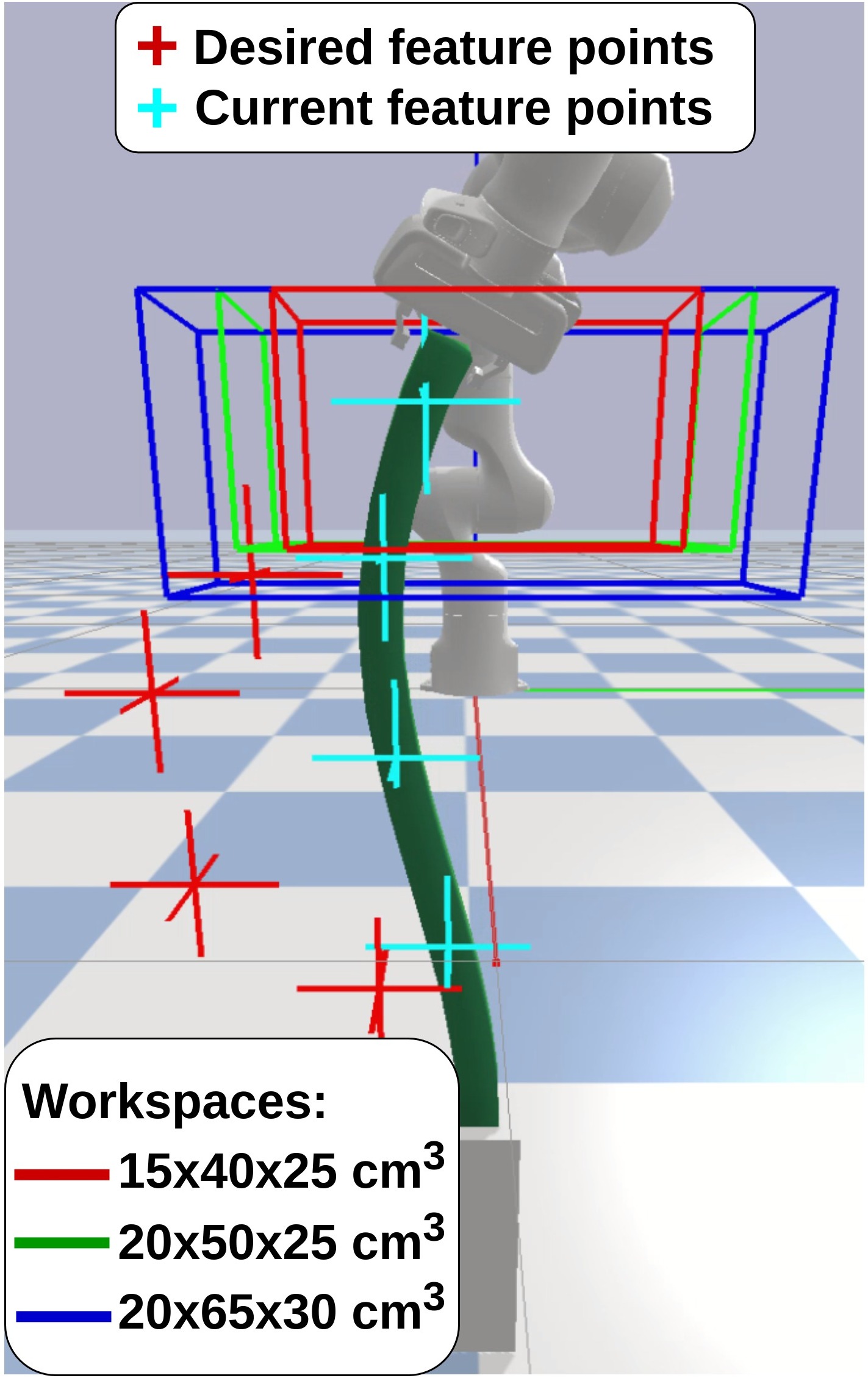}
         \label{box_fig_a}
     \end{subfigure}
     \hfill
     \begin{subfigure}[b]{0.2\textwidth}
         \includegraphics[height=5cm]{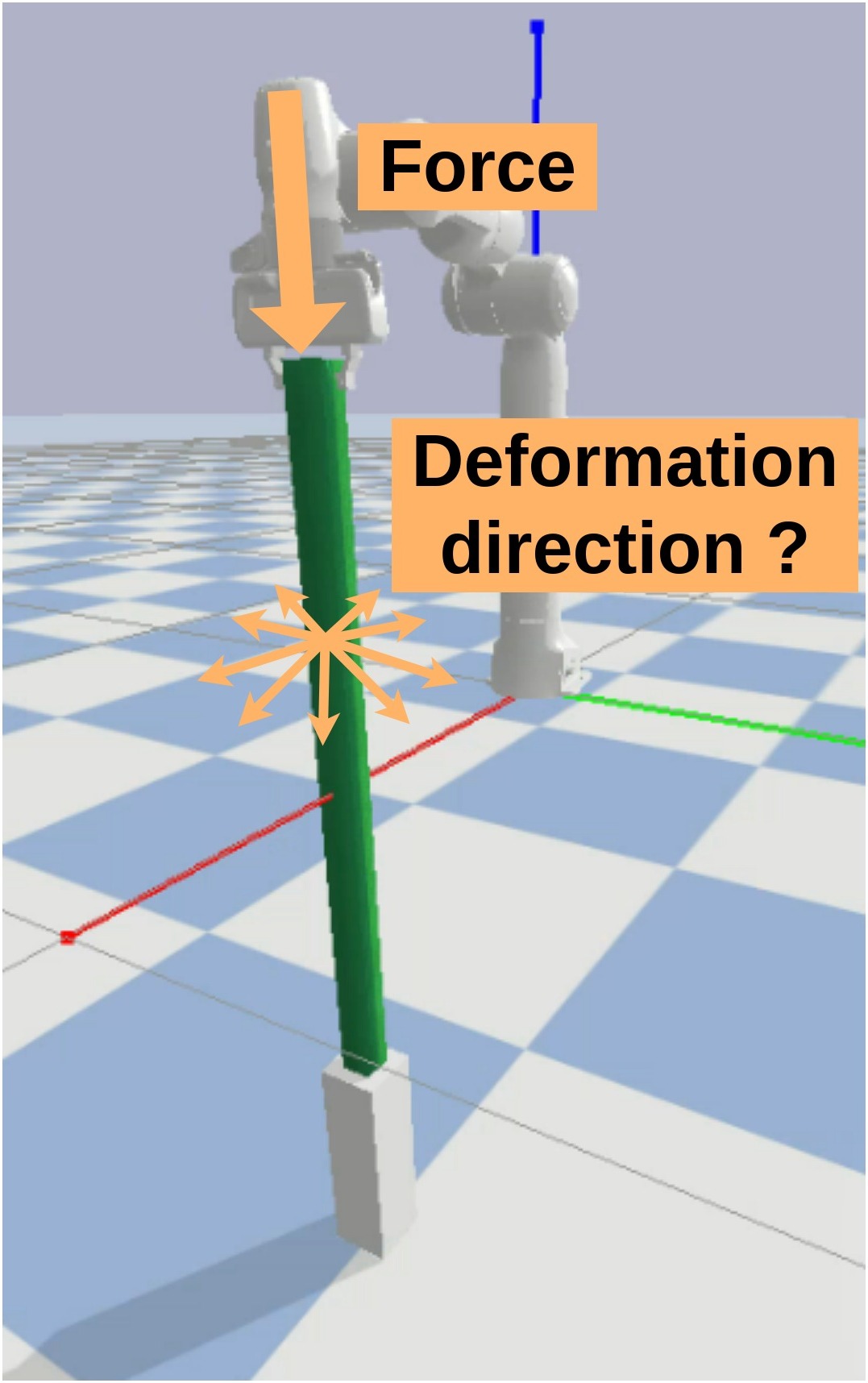}
         \label{box_fig_b}
     \end{subfigure}
     \hfill
     \vspace{-5pt}
    \caption{\Update \small Left: DLO manipulation in a simulated environment considering different workspaces. These were used to create different  deformation datasets to evaluate MultiAC6 (cf. Section~\ref{sec_experiments}). Right: Overview of a singular configuration for which the deformation cannot be predicted, which leads to a sim-to-real gap. \Done \squeezeup \squeezeup \squeezeup} 
    \label{Boxes}
\end{figure} 


\begin{figure*}[ht!]
\centering
\includegraphics[width=.85\textwidth]{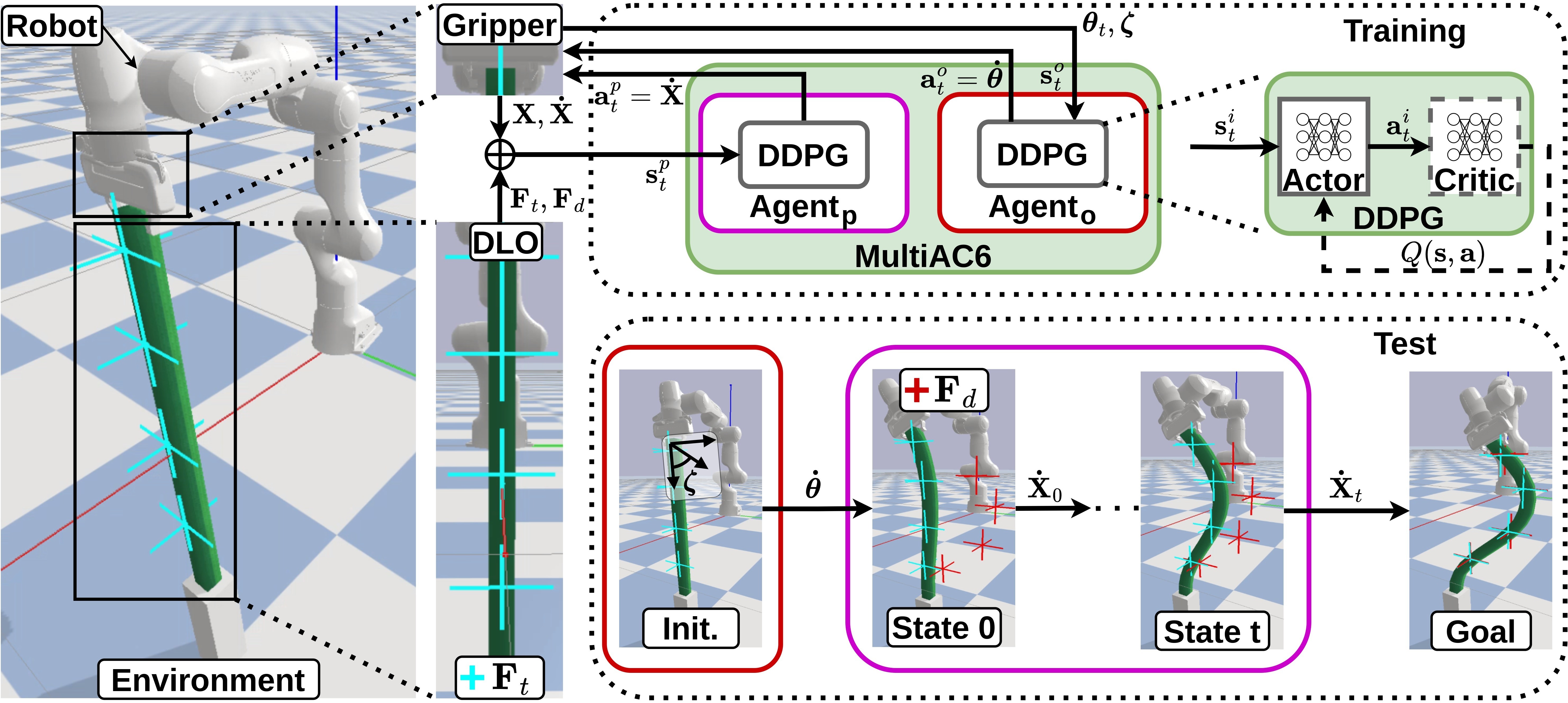}
\vspace{-0.12cm}
\caption{\small Overview of MultiAC6 framework for DLO manipulation. MultiAC6 decomposes the robot action space using two agents. $\text{Agent}_o$ orients the DLO tip towards $\zeta$, then $\text{Agent}_p$ positions the gripper to reach the desired deformation.}
\label{ac6_framework}
\vspace{-0.5cm}
\end{figure*} 

Let us consider the 3D manipulation of DLOs using a single-arm robot. In this configuration, we assume that a robot grasps one extremity of a DLO. The second extremity of this DLO is fixed to the ground. The DLO is long ($>$~60~cm) and is assumed to be elastic. An elastic deformation implies that the DLOs return to their original shape once the deformation force is no longer applied \cite{Sanchez2018}. The goal is then to control the pose of the robot gripper to shape the DLO with a desired deformation. The DLO shape is tracked in real-time by a set of feature points defined by their 3D positions $(x, y, z)$. \UpdateNew We assume that the feature points can be tracked accurately in real-time with a vision-based algorithm. \Done 

Therefore, the manipulation task consists in moving the gripper so that the DLO feature points reach target positions representing a desired shape (see Figure \ref{Boxes}). \Update To achieve a particular deformation, the robot gripper should follow a specific trajectory. Indeed, knowing the gripper final pose is not sufficient to guarantee the achieved deformation accuracy. There is an ambiguity related to the gripper configuration: a unique gripper pose may correspond to very different DLO deformations. Given this background, this work focuses specifically on achieving large deformations for objects with large strains (plants, tubes, etc.). Our approach can generate a suitable trajectory without needing online fine-tuning based on DLO deformation testing \cite{Reza2022, Chi2022}. This can help in avoiding damage to the DLO. \Done 
We quantify the magnitude of a DLO deformation as the maximum among the distances between every feature point's initial and target position.  Using the results in existing works as a criterion (as done in \cite{Yu2023}) we define large deformations as those that exceed 15 cm in real-world settings.

Similarly to many DRL frameworks, MultiAC6 is trained in a simulator, for safer interaction and shorter training time \cite{Zhao2020}. Unfortunately, the transfer of trained policies to real-world settings generally does not work well due to the sim-to-real gap \cite{Daniel2022}. The sim-to-real gap is caused by the difficulty of synthesizing realistic interactions in simulation (due to under-modeling, wrong/approximated parameters, model discrepancies, etc.). For DLOs, the sim-to-real gap cannot be solved using classical sim-to-real transfer techniques which mainly address perception \cite{Zhao2020}. Simulated DLOs differ significantly from real DLOs. First, mechanical parameters (Young's modulus, Poisson coefficient, mass, friction, etc.) are only valid for one instance of a DLO. Second, real DLOs may be elastoplastic \cite{Sanchez2018} and partially maintain deformations. \Update Finally, some simulated deformations do not match the real ones for the same action. This is the case for singular positions of DLOs \cite{Bretl2014, Borum2014}, as illustrated in Figure~\ref{Boxes}. In such a configuration, DLOs are at equilibrium, but unstable in the sense that any slight gripper motion leads to unpredictable deformations. This occurs, for example, when a vertical force is exerted on a straight DLO. Such a singularity is rarely addressed in the literature. The authors of \cite{Navarro2016} acknowledged this singularity as a limitation of their method: the singularity occurs for objects with very small curvature. In the case of analytical methods, the Jacobian matrix is singular and the model becomes unstable. \Done These singular deformations cannot be replicated in simulation. This discrepancy between real and simulated deformations violates the Markovian observability property \cite{Sutton1998} of the DRL methods. Consequently, the policies learned in simulation are no longer valid in real-world settings.

Achieving such large DLO deformations in 3D presents multiple challenges. To address these, we propose a new DRL framework, described in the next sections, based on DDPG.  \squeezeup

\Update

\section{Background on DDPG}
The DDPG algorithm is an off-policy actor-critic method used to deal with continuous action spaces \cite{Lillicrap2015DDPG}. Considering the continuous state space $\mathcal{S}$ and action space $\mathcal{A}$, the DDPG agent aims to learn the optimal policy $\pi^{*}: \mathcal{S} \longrightarrow \mathcal{A}$. The learning process involves the acquisition of a Q-function and a policy \cite{Jangir2020ICRA}. For this purpose, an actor, also known as the policy network, takes the current state $\mathbf{s}_t$ as input and generates the optimal action $\mathbf{a}_t$ as output. Simultaneously, a critic, called the Q-function network, assesses the optimality of the action chosen $\mathbf{a}_t$ in the state $\mathbf{s}_t$ by assigning Q-values $Q_t(\mathbf{s}_t, \mathbf{a}_t)$ to the state-action pair ($\mathbf{s}_t$, $\mathbf{a}_t$).

The actor and critic network are trained from data stored in a replay buffer. This replay buffer is filled with transitions. A transition $T$ is composed of the action $\mathbf{a}_t$ predicted by the actor, the state $\mathbf{s}_t$, the next state $\mathbf{s}^{\prime}_t$ after applying $\mathbf{a}_t$, and the reward $r_t$ obtained for ($\mathbf{s}_t$, $\mathbf{a}_t$). Actor and critic networks are trained when the replay buffer contains enough data $\ge N$ to extract a batch of non-sequential transitions (see Table~\ref{ddpg_parameters}). These transitions are selected randomly to guarantee that the data are independent and identically distributed.

Utilizing batches and allowing agents to learn from previous experiences accelerates the learning process while removing unwanted temporal correlations \cite{Liu2018AACCCC}. In fact, the critic network aims to minimize the error between the predicted Q-values $Q(\mathbf{s}, \mathbf{a})$ and the Q-values calculated using the Bellman equation \cite{Nachum2017ANIPS} $Q_B(\mathbf{s}, \mathbf{a}) = r + \gamma \times Q^{\prime}(\mathbf{s}^{\prime},\mathbf{a}^{\prime})$, where $\gamma$ is the discount factor. More specifically, the critic network is optimized by minimizing the mean square error (MSE) between $Q_B(\mathbf{s},\mathbf{a})$ and $Q(\mathbf{s} ,\mathbf{a})$. Given a batch size $N$ sampled from the replay buffer, the critic loss $\ell_c$ becomes:
\begin{equation}
   \ell_c = \frac{\sum_{n=1}^{N}({Q_B}_n(\mathbf{s}_n,\mathbf{a}_n)-{Q_n(\mathbf{s}_n,\mathbf{a}_n)})^{2}} {N}.
\end{equation}

The actor network predicts actions that maximize the Q-values. Therefore, the actor network is optimized by minimizing the negative Q-value. The policy loss $\ell_p$ is calculated by averaging $Q(\mathbf{s},\mathbf{a})$ \cite{Lillicrap2015DDPG}: \squeezeup
\begin{equation}\label{MSE_actor}
    \ell_p = -\overline{Q(\mathbf{s},\mathbf{a})} = - \frac{\sum_{n=1}^{N} Q_n(\mathbf{s}_n,\mathbf{a}_n)}{N}.
\end{equation}
\Done
\squeezeup \squeezeup
\section{Method}
\Update

\subsection{Action and State Spaces}
In the MultiAC6 framework (see Figure~\ref{ac6_framework}), the action space of a robot is divided between two agents. Each agent within MultiAC6 is a DDPG agent. 
In this setup, the goal is to control the gripper pose $\mathcal{P}$. Let us define $\mathcal{P} = (\mathbf{X},\boldsymbol{\theta})$ with $\mathbf{X} = (x,y,z)$ the position and $\boldsymbol{\theta} = (\theta_x,\theta_y,\theta_z)$ the orientation of the gripper in the world frame. From this notation, let us define a position agent ($\text{Agent}_p$) that actuates the gripper translation velocity $\mathbf{\dot X}$, and an orientation agent ($\text{Agent}_o$) that actuates the gripper angular velocity $\boldsymbol{\dot\theta}$. In this framework, the robot deforms a DLO to minimize the error between the current feature points $\mathbf{F}$ and the desired feature points $\mathbf{F}_d$. The division of the action space between two agents is also translated into the way the task is performed. First, $\text{Agent}_o$ orients the gripper and then $\text{Agent}_p$ positions the gripper so that the desired deformation is achieved.

In a timestep $t$ and considering the continuous state space $\mathcal{S}$ and action space $\mathcal{A}$, the $\text{Agent}_p$ action $\mathbf{a}^p_t \in \mathcal{A}^p$ is $\mathbf{\dot X}_t \in \mathbb{R}^3$. The $\text{Agent}_p$ state $\mathbf{s}^p_t$ consists of the current position $\mathbf{X}_t$ and the translation velocity $\mathbf{\dot X}_t$ of the gripper, and the current and desired feature points. Hence, $\mathbf{s}^p_t \in \mathcal{S}^p$  is $(\mathbf{X}, \mathbf{\dot X}, \mathbf{F}, \mathbf{F}_d)_t \in \mathbb{R}^{6+6m}$, with $m$ the number of selected feature points. 

For $\text{Agent}_o$, its action $\mathbf{a}^o_t \in \mathcal{A}^o$ is defined as $\boldsymbol{\dot \theta}_t \in \mathbb{R}^3$. The state of $\text{Agent}_o$ $\mathbf{s}^o_t$ consists of the current gripper orientation $\boldsymbol{\theta}_t$, the desired DLO tip orientation $\boldsymbol{\zeta} = (\zeta_x, \zeta_y, \zeta_z)$, and the desired feature points. Hence, $\mathbf{s}^o_t \in \mathcal{S}^o$ is $(\boldsymbol{\theta}, \boldsymbol{\zeta}, \mathbf{F}_d)_t \in \mathbb{R}^{6+3m}$. The $\text{Agent}_o$ state is designed in a similar way as many DRL-based manipulation tasks \cite{JCBAO2023, Marzari2021HLC, LCHEN2022} specifying the desired goals in the state vector.

\subsection{MultiAC6 action space decomposition} \label{MultiAC6_training}
\subsubsection{Principle}
The proposed action space decomposition provides a straightforward but still efficient way to bridge the sim-to-real gap for DLO manipulation. For this purpose, a specific decoupled training strategy is proposed as follows. In our settings, $\text{Agent}_o$ is trained to achieve a given desired DLO tip orientation $\boldsymbol{\zeta}$. This desired orientation $\boldsymbol{\zeta}$ is hand-crafted (only for training) and is known to lead to the desired DLO deformation. Therefore, this agent is not trained with the simulator. Indeed, the $\text{Agent}_o$ state ($\boldsymbol{\theta}, \boldsymbol{\zeta}, \mathbf{F}_d$) is independent of the DLO deformation represented by the feature points $\mathbf{F}_t$. With the assumption that the DLO tip is locally rigid, the gripper orientation $\boldsymbol{\theta}_t$ can be obtained by integrating $\boldsymbol{\dot \theta}_t $. It is worth noting that $\boldsymbol{\zeta}$ is defined to avoid singular configurations of the DLO. As a direct benefit, the sim-to-real gap can be avoided for $\text{Agent}_o$. In parallel, from the desired orientation of the DLO tip, $\text{Agent}_p$ is trained to control the translation velocity of the gripper to deform the DLO into the desired shape. Given that $\text{Agent}_p$ always starts from a DLO oriented with $\boldsymbol{\zeta}$, the sim-to-real gap related to singular DLO configurations can also be avoided.


Each of the agents is trained separately to avoid error accumulation. This strategy has been used in \cite{Marzari2021HLC} for a pick-and-place task where it has been proven to outperform a sequential training strategy. Since MultiAC6 agents are trained separately, issues of non-stationary environments \cite{Wang2021} are avoided. Such issues occur when both agents update the environment simultaneously. $\text{Agent}_p$ and $\text{Agent}_o$ would not be able to correctly map states to actions. Therefore, learning an optimal policy would be more challenging. 

When both agents are trained, the manipulation task is solved in several steps.  First, $\text{Agent}_o$ orients the DLO tip towards $\boldsymbol{\zeta}$. Thereafter, $\text{Agent}_p$ positions the gripper, so that the feature points $\mathbf{F}_t$ reach the target points $\mathbf{F}_d$.

\subsubsection{Theoretical reasoning}
Although DRL approaches usually only control the position of the gripper, it is more intuitive and natural to also actuate the gripper orientation. A 6 DOF-gripper is less restricted and can subsequently achieve more complex deformations than a 3 DOF-gripper. Furthermore, as mentioned previously, singular  configurations can be avoided with a proper orientation of the DLO tip.
Unfortunately, using more DOFs leads to the well-known curse of dimensionality inherent in DRL approaches: the action space grows exponentially with the number of controlled DOFs. It becomes more difficult to find an optimal policy to achieve the desired DLO deformations.
To mitigate this issue, our proposed action space decomposition framework combines the advantages of a 6 DOF control of the gripper with the benefits of a limited action space. Indeed, by decoupling the gripper control over two agents, each of them only explores a limited action space, allowing them to find useful learning signals to achieve their respective task.

\subsection{Optimization framework} \label{ddpg}
\subsubsection{Learning parallelization}
MultiAC6 uses the learning parallelization technique introduced for the A3C (asynchronous advantage actor-critic) algorithm \cite{Mnih2016A3C}. The principle is to run multiple agents simultaneously in parallel on different environments. With this approach, more data can be collected for a given time period. For off-policy algorithms such as DDPG, the replay buffer is filled faster. Furthermore, since agent environments and actions are not correlated, transitions containing more diverse state-action pairs can be collected in the replay buffer. Therefore, learning parallelization decreases training time while yielding better results, as shown in \cite{Daniel2022}.


\subsubsection{Reward function} \label{reward_def}
The reward function controls the optimization of the agent action selection policy \cite{Zakaria2021Frontiers}. For $\text{Agent}_p$, the reward function ${r^p_1}_t$ is defined as the maximum error. ${r^p_1}_t$ is computed as the negative of the maximum Euclidean distance $D_t$ between the current feature points and the desired feature points: \squeezeup
\begin{equation} \label{max_reward} 
    {r^p_1}_t  = -\max({D_t(\mathbf{F}_t,\mathbf{F}_d)}).  
\end{equation}
For $\text{Agent}_o$, the reward function $r^o_t$ is defined as the root-mean-square error (RMSE) between the current Euler orientation (roll, pitch, yaw angles) of the gripper and the desired DLO tip orientation: \squeezeup
\begin{equation} \label{MSE} 
    r^o_t = - \text{RMSE}(\boldsymbol{\theta}_t,\boldsymbol{\zeta}).  
\end{equation}

\squeezeup 

\Done

\section{Experiments} \label{sec_experiments}

\subsection{Simulation setup}
\subsubsection{Environment configuration}
As mentioned in previous sections, a simulator is required to train the DDPG agents. For this purpose, PyBullet, the Python version of Bullet \cite{Coumans2021Pybullet}, was used as the simulator physics engine. The simulated environment consisted of a 7-DOF Franka Emika Panda robot and a DLO of dimension 5x5x103 cm$^3$. The DLO deformations were modeled using FEM. A unique DLO model was defined with a 3D tetrahedral mesh comprising 70 nodes, 104 tetrahedrons, 241 links and 136 faces. This DLO was characterized by a Young’s modulus of 2.5 MPa, a Poisson coefficient of 0.3, a mass of 0.2 Kg, a damping ratio of 0.01, and a friction coefficient of 0.5. In the simulator, the current feature points $\mathbf{F}_t$ and the desired feature points $\mathbf{F}_d$ were defined using the positions of some mesh nodes. Four mesh nodes ($m=4$) were selected all along the DLO (cf. Figure~\ref{Boxes}). This number is enough to characterize the DLO shape and works well in practice \cite{Daniel2022}.

\begin{table}[t!] \vspace{-0.4cm}
\caption{\small DDPG parameters}
\vspace{-0.5cm}
\label{ddpg_parameters}
\begin{center}
\begin{tabular}{c|c} 
\toprule
\textbf{Parameter} & \textbf{Value} \\
\hline
Nb. layers & 3 \\
Hidden size & 256 \\
$\alpha_A$ & 0.0001 \\
$\alpha_C$ & 0.001 \\
Replay buffer & 50,000 \\
Batch size $N$ & 128 \\
$\gamma$ & 0.99 \\
\bottomrule
\end{tabular} \vspace{-0.2cm}
\end{center}
\squeezeup \squeezeup
\end{table}

\subsubsection{Datasets}
Three datasets of deformations were created to evaluate MultiAC6. Each of these datasets was collected in workspaces of different dimensions, as illustrated in Figure~\ref{Boxes}. The workspaces are defined as follows:
\begin{itemize}
    \item A small 15x40x25 cm$^3$ workspace which is used to collect both the training/seen test dataset.
    \item A medium 20x50x25 cm$^3$ workspace which is used to collect the unseen test dataset.
    \item A large 20x65x30 cm$^3$ workspace which is used to collect the large unseen test dataset.
\end{itemize}
Each dataset contained 1000 deformations defined by $\mathbf{F}_d$ and $\boldsymbol{\zeta}$. \Update The unseen datasets were excluded from the training phase to assess how well MultiAC6 could handle unseen samples. \Done It is worth noting that the large unseen dataset corresponded to the full robot workspace. Deformations are collected within each workspace by moving the gripper to a random pose. 

\begin{table}[tb!]
\caption{\small \UpdateNew Simulation results for MultiAC6$^*$ for different reward functions. With AE (the standard deviation $\sigma$)  (cm), and ME (cm).} 
\vspace{-0.5cm}
\label{reward_comparison}
\begin{center}
\scalebox{0.95}{
\begin{tabular}{p{1cm}|C{0.3cm}|C{0.5cm}  C{1.1cm}  C{0.5cm} | C{0.5cm} C{1.1cm} C{0.5cm}}
\toprule
\textbf{Reward} & $\boldsymbol{\delta_p}$ & \multicolumn{6}{c}{\textbf{Test Seen}} \\
\cline{3-8}
\textbf{Function} &  & \multicolumn{3}{c|}{\textbf{80 episodes}}  & \multicolumn{3}{c}{\textbf{100 episodes}} \\
\cline{3-8}
 & {} & \textbf{SR$\uparrow$} & \textbf{AE$\downarrow$} & \textbf{ME$\downarrow$} & \textbf{SR$\uparrow$} & \textbf{AE$\downarrow$} & \textbf{ME$\downarrow$} \\
\hline
Maximum           & 5  & \textbf{1.0}  & \textbf{3.38}(0.96) & 1.09  & \textbf{1.0}  & \textbf{3.04}(1.09) & 1.01  \\
error ${r^p_1}$   & 3  & \textbf{0.98} & \textbf{2.34}(0.57) & \textbf{0.40}   & 0.99          & \textbf{2.01}(0.64) & 0.67  \\
\hline
Mean               & 5  & 0.93 & 4.11(1.32) & \textbf{0.94}  & \textbf{1.0}  & 3.40(0.99)  & \textbf{0.66}   \\
error ${r^p_2}$    & 3  & 0.52 & 3.54(1.63) & 0.66   & \textbf{1.0}  &  2.04(0.61) & \textbf{0.20}   \\
\hline
DTW                & 5  & 0.86 & 4.44(1.49) & 1.04 & 0.94          & 3.86(1.20)  & 1.22  \\
${r^p_3}$          & 3  & 0.52 & 3.91(1.77) & 1.02 & 0.76          & 3.10(1.32)  & 0.99  \\
\bottomrule
\end{tabular}}
\end{center}
\end{table}

\begin{table}[tb!]
\vspace{-0.6cm}
\begin{center}
\scalebox{0.95}{
\begin{tabular}{p{1cm}|C{0.3cm}|C{0.5cm}  C{1.1cm}  C{0.5cm} | C{0.5cm} C{1.1cm} C{0.5cm}}
\toprule
\textbf{Reward} & $\boldsymbol{\delta_p}$ & \multicolumn{6}{c}{\textbf{Test Unseen}} \\
\cline{3-8}
\textbf{Function} &  & \multicolumn{3}{c|}{\textbf{80 episodes}}  & \multicolumn{3}{c}{\textbf{100 episodes}} \\
\cline{3-8}
 & {} & \textbf{SR$\uparrow$} & \textbf{AE$\downarrow$} & \textbf{ME$\downarrow$} & \textbf{SR$\uparrow$} & \textbf{AE$\downarrow$} & \textbf{ME$\downarrow$} \\
\hline
Maximum          & 5  & \textbf{0.99} & \textbf{3.61}(0.90) & \textbf{0.78} & \textbf{1.0} & \textbf{3.23}(1.03) & \textbf{0.47}  \\
error ${r^p_1}$  & 3  & \textbf{0.89} & \textbf{2.55}(0.78) & \textbf{0.23} & \textbf{0.97}& \textbf{2.10}(0.73) & 0.49  \\
\hline
Mean            & 5  & 0.89 & 4.32(1.31) & 1.51 & \textbf{1.0} & 3.54(0.95) & 1.14 \\
error ${r^p_2}$ & 3  & 0.47 & 3.83(1.68) & 0.69 & \textbf{0.97} & 2.21(0.67) & \textbf{0.34} \\
\hline
DTW             & 5 & 0.85  & 4.51(1.68) & 1.11 & 0.92         & 4.13(1.60) & 0.93  \\
${r^p_3}$       & 3 & 0.44  & 4.22(1.97) & 1.16 & 0.65         & 3.52(1.85)  & 0.97 \\
\bottomrule
\end{tabular}}
\end{center}
\end{table}

\begin{table}[tb!]
\vspace{-0.6cm}
\begin{center}
\scalebox{0.95}{
\begin{tabular}{p{1cm}|C{0.3cm}|C{0.5cm}  C{1.1cm}  C{0.5cm} | C{0.5cm} C{1.1cm} C{0.5cm}}
\toprule
\textbf{Reward} & $\boldsymbol{\delta_p}$ & \multicolumn{6}{c}{\textbf{Test Large Unseen}} \\
\cline{3-8}
\textbf{Function} &  & \multicolumn{3}{c|}{\textbf{80 episodes}}  & \multicolumn{3}{c}{\textbf{100 episodes}} \\
\cline{3-8}
 & {} & \textbf{SR$\uparrow$} & \textbf{AE$\downarrow$} & \textbf{ME$\downarrow$} & \textbf{SR$\uparrow$} & \textbf{AE$\downarrow$} & \textbf{ME$\downarrow$} \\
\hline
Maximum         & 5  & \textbf{0.88} & \textbf{4.62}(3.79)  & \textbf{0.79} & \textbf{0.96} & \textbf{3.99}(3.87)  & 0.97 \\ 
error ${r^p_1}$ & 3  & \textbf{0.59} & \textbf{3.91}(3.98)  & 0.79 & \textbf{0.89} & \textbf{2.96}(3.98)  & \textbf{0.44} \\
\hline
Mean               & 5  & 0.70 & 5.36(2.42)  & 1.36 & 0.94 & 4.03(0.22)  & \textbf{0.54} \\
error ${r^p_2}$    & 3  & 0.31 & 5.05(2.71)  & \textbf{0.34} & 0.79 & 3.08(2.43)  & 0.54 \\
\hline
DTW           & 5  & 0.62 & 7.11(9.59)  & 1.25 & 0.73 & 6.63(9.18)  & 1.13 \\
${r^p_3}$     & 3  & 0.25 & 7.00(9.66)  & 1.13 & 0.33 & 6.41(9.31)  & 1.05 \\
\bottomrule
\end{tabular}}\vspace{-0.85cm}
\end{center}
\end{table}\Done

\begin{table*}[tb]
\caption{\small AC3, AC6, and MultiAC6 simulation results for the test seen, unseen, and large unseen datasets. With AE $\pm$ the standard deviation $\sigma$ in cm, and ME in cm.}
\label{agents_comparison_simulation}
\vspace{-0.5cm}
\begin{center}
\begin{tabular}{l|c|*3c|*3c|*3c}
\toprule
\textbf{Method} & $\boldsymbol{\delta_p}$ & \multicolumn{3}{c|}{\textbf{Test Seen}} & \multicolumn{3}{c|}{\textbf{Test Unseen}} & \multicolumn{3}{c}{\textbf{Test Large Unseen}}\\
\cline{3-11}
{} & {} & \textbf{SR $\uparrow$} & \textbf{AE $\downarrow$} & \textbf{ME $\downarrow$} & \textbf{SR $\uparrow$} & \textbf{AE $\downarrow$} & \textbf{ME $\downarrow$} & \textbf{SR $\uparrow$} & \textbf{AE $\downarrow$} & \textbf{ME $\downarrow$}\\
\hline
AC3 \cite{Daniel2022} & 5  & 0.64  & 4.83 $\pm$ 1.22 & 1.78  & 0.59 & 9.04 $\pm$ 8.60 & \textbf{1.34}  & 0.49 & 12.17 $\pm$ 11.37 & 1.66 \\
                      & 3  & 0.26 &  4.45 $\pm$ 1.57 & 1.55  & 0.33 & 8.56 $\pm$ 8.90 & 1.08  & 0.30 & 11.75 $\pm$ 11.68 & 1.12 \\
\hline
AC6                   & 5  & \textbf{1.0}  & 4.40 $\pm$ 0.44  & 2.08  & 0.60 & 9.50 $\pm$ 7.76 & 2.04  & 0.47 & 12.39 $\pm$ 10.12  & 1.48 \\
                      & 3  & \textbf{1.0}  & 2.61 $\pm$ 0.36  & 1.26  & 0.54 & 8.55 $\pm$ 8.40 & 1.47  & 0.39 & 11.76 $\pm$ 10.63  & 1.20 \\
\hline
MultiAC6 (ours)      & 5  & \textbf{1.0}  & \textbf{3.02} $\pm$ 1.07   & \textbf{0.98}  & \textbf{1.0} & \textbf{3.23} $\pm$ 1.02     & \textbf{1.34}  & \textbf{0.96} & \textbf{3.99} $\pm$ 3.86      & \textbf{0.99} \\
                      & 3  & 0.99          & \textbf{2.01} $\pm$ 0.62   & \textbf{0.62}  & \textbf{0.97}          & \textbf{2.09} $\pm$ 0.71    & \textbf{0.52}  & \textbf{0.89} & \textbf{2.95} $\pm$ 3.98      & \textbf{0.47} \\
\bottomrule
\end{tabular}\squeezeup\squeezeup \squeezeup
\end{center}
\end{table*}

\subsection{Training configuration}
\subsubsection{DDPG parameters} 
\Update The DDPG parameters were obtained empirically (see Table~\ref{ddpg_parameters}). \Done The actor and critic networks consisted of three fully connected hidden layers of dimension 256 with a rectified linear unit (ReLU) activation function. The actor output $\mathbf{a}_t$ was passed through a Tanh activation function. For exploration purposes, Ornstein-Uhlenbeck noise was added to the action $\mathbf{a}_t$, as described in \cite{Lillicrap2015DDPG}. The network gradients were updated with the ADAM optimizer. The learning rate was set to $\alpha_A = 0.0001$ for the actor and $\alpha_C = 0.001$ for the critic. A batch size of $N = 128$ transitions was randomly sampled from a 50000-size replay buffer. Finally, the discount factor was set to a constant value ($\gamma = 0.99$).

\subsubsection{Training parameters} \label{multi_reward}
$\text{Agent}_p$ was trained with 32 parallel agents for 100 episodes of 300 steps. In this configuration, a manipulation task was considered successful when the maximum error (as defined in Section~\ref{reward_def}) was below a threshold $\delta_p$ set at 5 cm. \Update This threshold is generally sufficient for applications such as manipulating plants. From this, we could define the success rate (SR). \Done Similarly, for $\text{Agent}_o$, 32 agents were trained in parallel for 60 episodes of 100 steps. The training dataset was used to sample the desired mesh nodes $\mathbf{F}_d$. The angular error threshold $\delta_o$ was set to 3° (or 0.0524 rad). Both $\text{Agent}_p$ and $\text{Agent}_o$ were trained on supercomputers with 64 GB memory and Intel Xeon E5-2698 v4 2.20 GHz processors at the UCA University Mesocentre. The average training time was two and a half days, mainly due to the slowness of the FEM computation. \UpdateNew The training time can be reduced by using more powerful computers or optimized simulators such as Isaac Gym \cite{Makoviychuk2021}.
\Done

\subsection{Simulation results}
Several experiments were conducted in simulation to \textit{(i)} assess the performance of the $\text{Agent}_p$ reward function, and \textit{(ii)} evaluate the MultiAC6 framework. All results were obtained for 1000 desired random goals sampled from the seen, unseen, and large unseen datasets. \Update Several evaluation metrics were used, which are SR, average error (AE), and minimum final error (ME). \Done

\subsubsection{Reward function evaluation}
A first evaluation consisted in assessing the performance of our proposed reward function. \Update For this purpose, the maximum error reward function was compared to a mean error and a dynamic time warping (DTW) reward function \cite{Berndt1994}. The DTW reward computes the similarity between two point sets (DLO feature points). \Done The mean error reward function was calculated as the negative average Euclidean distance $D_t$ between the current feature points $\mathbf{F}_t$ and the desired feature points $\mathbf{F}_d$: \vspace{-0.1cm}
\begin{equation} \label{mean_reward}
    {r^p_2}_t = -\overline{D_t(\mathbf{F}_t,\mathbf{F}_d)} = - \frac{\sum_{j=1}^{m} D_t(\mathbf{F}_{t_j},\mathbf{F}_{d_j})}{m}.
\end{equation}

The DTW reward function was used for measuring the similarity between the current feature points $\mathbf{F}_t$ and the desired feature points $\mathbf{F}_d$: \Update \squeezeup
\begin{equation} \label{dtw_reward}
    {r^p_3}_t = - \text{DTW}(\mathbf{F}_t,\mathbf{F}_d) = - \sum_{j=1}^{m} D_t(\mathbf{F}_{t_j},\mathbf{F}_{d_j}).
\end{equation}
\Done \squeezeup

To capture only the effect of the reward functions, only $\text{Agent}_p$ was evaluated with the initial hand-crafted DLO tip orientation. This framework was denoted MultiAC6$^*$. \UpdateNew As shown in Table~\ref{reward_comparison}, the maximum error reward function performed overall well. With 80 episodes, our proposed reward function always performed the best with large differences in success rates compared to the DTW or the mean error reward. With 100 training episodes, 89\% of the deformations were successfully performed under the most challenging condition (large unseen with $\delta_p =$ 3 cm) for the maximum error reward. In comparison, the mean error reward function had a success rate of 79\% while the DTW reward function only achieved 33\%. These results with 80 or 100 training episodes support the superiority of our proposed reward.  We believe that the maximum error reward performs better because it does not smooth the error as with the mean error reward. Furthermore, this reward is easier to maximize than the DTW error reward. For the following experiments, a maximum error reward was used with 100 training episodes. \Done

\begin{figure*}
    \centering
    \includegraphics[width=\textwidth]{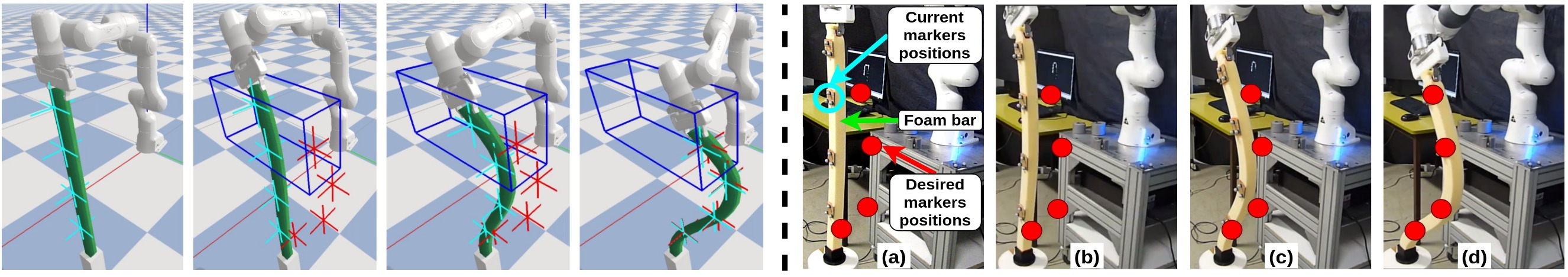}\vspace{-3pt}
    \caption{\small Deformation performed by MultiAC6 with a one-meter long foam bar and $\delta_p  = $ 5 cm. The initial configuration is given in (a), then in (b) $\text{Agent}_o$ orients the gripper, and finally in (c)-(d), $\text{Agent}_p$ positions the gripper to reach the desired deformation.  \squeezeup  \squeezeup}
    \label{real_deformation}
\end{figure*}

\subsubsection{MultiAC6 evaluation}
The MultiAC6 framework was then compared with different approaches. In particular, MultiAC6 was compared with single-agent frameworks for controlling the 6 DOF (AC6) or 3 DOF (AC3 \cite{Daniel2022}) of the robot gripper. AC6 is a single-agent framework that directly outputs both translation and angular velocities. AC3 and AC6 were trained for 100 episodes of 300 steps under the same conditions and with the same parameters as MultiAC6. As shown in Table~\ref{agents_comparison_simulation}, AC3 performed poorly even with seen deformations. As initially assumed, controlling 3 DOF is not sufficient to achieve large 3D deformations. Differently, AC6 performed well only with seen deformations. The success rate dropped drastically for unseen datasets  (down to 39\%). This suggests that AC6 may not be able to perform well in real-world conditions. In contrast, our MultiAC6 framework achieved at least 89\% deformations even under the most challenging conditions (see Figure~\ref{real_deformation}). These results, which are consistent with \cite{Wang2021}, confirm the benefit of using the action space decomposition. Indeed, with MultiAC6, agents explore smaller state-action spaces than single-agent frameworks. 



Furthermore, on average, deformations are achieved with an accuracy between 2 and 3 cm. This accuracy can reach in the best-case scenarios 0.51 cm. These results obtained with datasets involving unseen deformations demonstrate the robustness of the MultiAC6 framework.

\vspace{-0.1cm}
\subsection{Experimental results}
For real-world experiments, we used a 7-DOF Franka Emika Panda robot to manipulate a long foam bar as illustrated in Figure~\ref{real_deformation}(a). Feature points $\mathbf{F}_t$ on the foam bar were defined by markers. These markers were tracked in real-time with a motion capture (MOCAP) system. For all experiments, the threshold $\delta_p$ was set to 5 cm for $\text{Agent}_p$ and $\delta_o$ was 3° for $\text{Agent}_o$. 


\begin{figure}[tb]
\centering
\includegraphics[width=6cm]{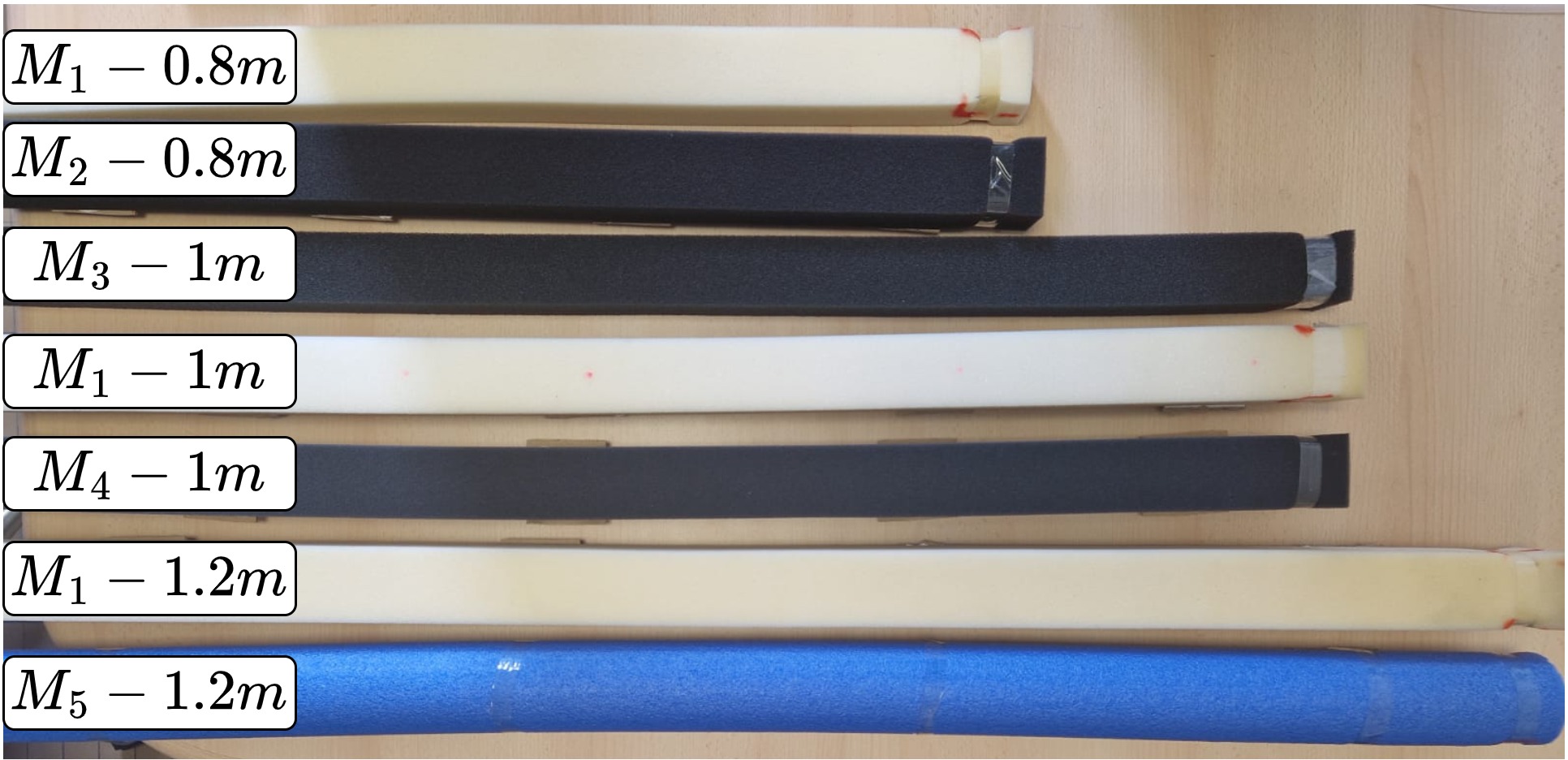}\vspace{-4pt}
\caption{\small Foam bars with different lengths and materials that have been used in the real experiments.}  
\label{foam_bars}
\end{figure} 

\begin{figure}[tb]\vspace{-0.4cm}
\centering
\includegraphics[width=6.5cm]{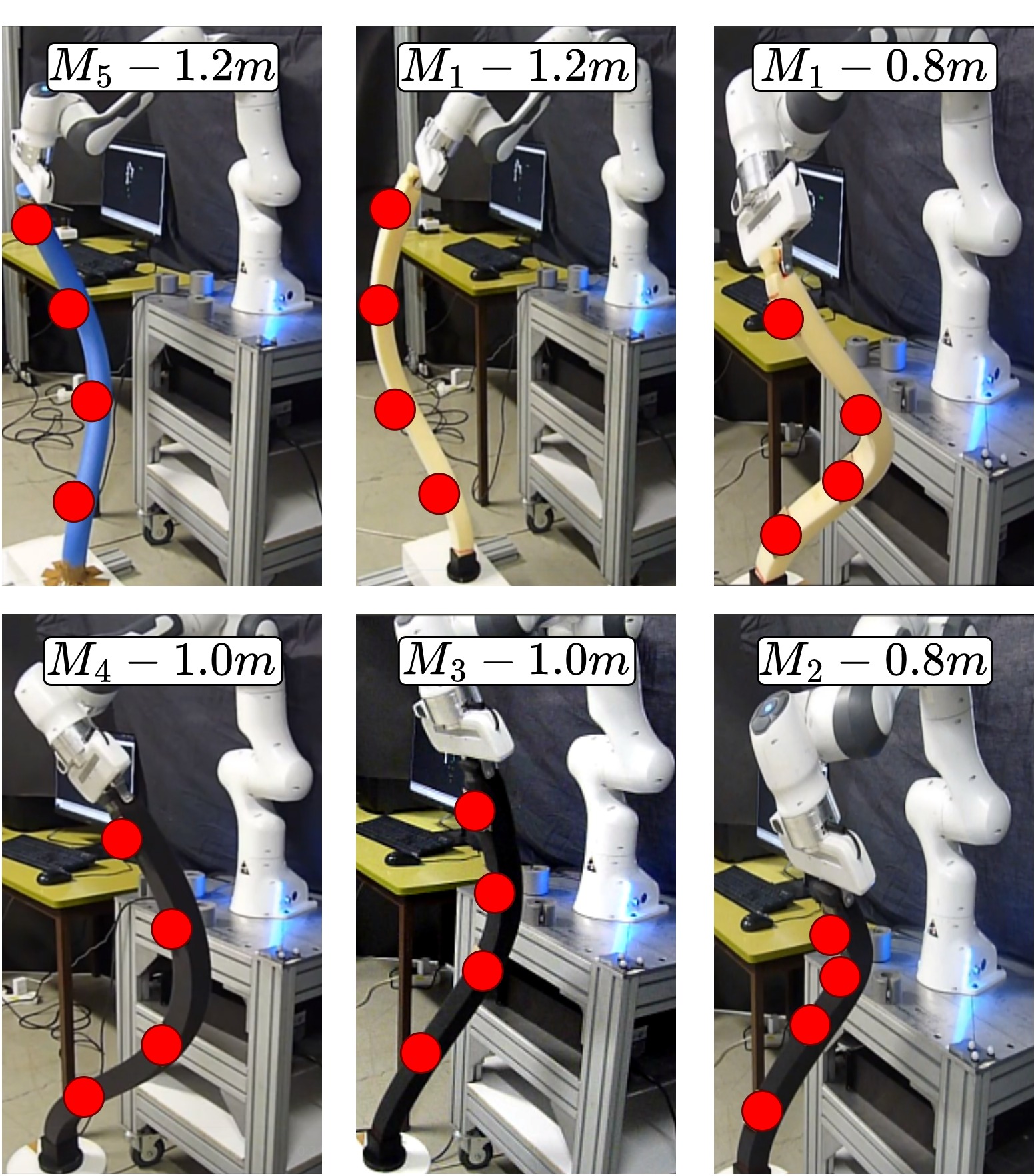}\vspace{-3pt}
\caption{\small Various deformations achieved by MultiAC6 with different foam bars.}
\label{all_deformations}\vspace{-0.8cm}
\end{figure}

\subsubsection{MultiAC6 real-world evaluation} \label{MultiAC6_real_tests}
The experimental results are presented in Table~\ref{agents_comparison_real}. These results were obtained using 30 samples of reachable desired deformations. The success rate of AC3 and AC6 was 7/30 and 9/30, respectively. In contrast, MultiAC6 achieved 29/30 (+66\% compared to AC6) deformations with an average error of 3.65 cm. As hypothesized from the simulation results, AC6 was not able to overcome the sim-to-real gap. By analyzing the results, we discovered that AC6 was heavily affected by the elastoplasticity of the foam bar (different from the initial elasticity assumption) as well as singular configurations. On the contrary, MultiAC6 was able to avoid singular configurations thanks to the decoupled training framework of $\text{Agent}_p$ and $\text{Agent}_o$ (see Section~\ref{MultiAC6_training}). With the additional benefit of the action space decomposition, MultiAC6 policies are more efficient and thus transferable to real-world settings.



\begin{table}[t]
\caption{\small AC3, AC6, and MultiAC6 results in real world with a one-meter long DLO.}
\label{agents_comparison_real}
\vspace{-0.5cm}
\begin{center}
\begin{tabular}{l|c|c|c} 
\toprule
\textbf{\shortstack{Method}} & \textbf{SR $\uparrow$} & $\boldsymbol{\Delta}$ & \textbf{AE $\pm \text{ } \boldsymbol{\sigma}$ (cm) $\downarrow$} \\
\hline
AC3 \cite{Daniel2022} & 7/30 & -73\% & 12.10 $\pm$ 8.73 \\
AC6 & 9/30 & -66\% & 11.46 $\pm$ 8.41 \\
MultiAC6$^*$ & 29/30 & 0\% & 3.66 $\pm$ 0.84 \\
MultiAC6 (ours) & \textbf{29/30} & \textemdash &  \textbf{3.65} $\pm$  0.86 \\
\bottomrule
\end{tabular} \vspace{-0.5cm}
\end{center}
\end{table}

\begin{table}[tb]
\caption{\small MultiAC6 real-world experiments results with different types of foam bars. With YM Young's modulus defined in (MPA), Stiffness defined in (N/mm), and Length defined in (m).}
\label{DLO_comparison}
\vspace{-0.5cm}
\begin{center}
\begin{tabular}{l|*3c|c|c} 
\toprule
 & \multicolumn{3}{c|}{\textbf{Foam bar parameters}} & \textbf{Success}  & \textbf{Initial deformation} \\
 \cline{2-4}
\textbf{Type} & \textbf{YM} & \textbf{Stiffness} & \textbf{Length} & \textbf{rate $\uparrow$} & \textbf{Max/Mean (cm)}  \\
\hline
      &    &  & 0.8 & 15/17 & 34.78/25.53 \\
$M_1$ & 0.10 & 4.8   & 1.0 & \textbf{17/17} & 35.61/28.86 \\
{}    & {}   & {}   & 1.2 & \textbf{17/17} & 40.54/27.41 \\
\hline
$M_2$ & 0.07 & 3.6  & 0.8 & \textbf{17/17} & 37.28/25.18\\
\hline
$M_3$ & 0.16 & 7.5  & 1.0 & 14/17 & 33.40/25.28 \\
\hline
$M_4$ & 0.05 & 2.8  & 1.0 & \textbf{17/17} & 37.74/27.14 \\
\hline
$M_5$ & 0.59 & 38.6 & 1.2 & 16/17 & 38.70/27.30 \\
\bottomrule
\end{tabular}\vspace{-0.5cm}
\end{center}
\end{table} 

\subsubsection{MultiAC6 robustness}
To test further the robustness of our approach, MultiAC6 was evaluated on seven foam bars (see Figure~\ref{foam_bars}) with different characteristics. These characteristics involved different lengths, materials, and stiffness, as presented in Table~\ref{DLO_comparison}. \Update We believe that these characteristics are relevant to capture MultiAC6 robustness to significantly different DLOs. The foam bars made of materials $M_1$ to $M_4$ were cubical (section = 5x5 cm$^2$). The foam bar made of $M_5$ was cylindrical (diameter = 6.5 cm). \Done The results in Table~\ref{DLO_comparison} were obtained from 17 samples of reachable desired deformations. The same MultiAC6 model as in Section~\ref{MultiAC6_real_tests} was used without additional training or online fine-tuning. MultiAC6 achieved 95\% of all deformations with very different types of foam bars. This result emphasizes the flexibility of our approach, which is particularly suitable for real-world applications (see Figure~\ref{all_deformations}). \Update We hypothesize that MultiAC6 can generalize well to different workspaces and materials, as agents mainly learn the dynamics of any DLO, and not the model of the DLO manipulated in simulation. \Done Furthermore, Table~\ref{agents_comparison_real} results showed that MultiAC6 was able to achieve large deformations (26 cm on average). Some configurations even exceeded 40 cm.

\subsection{Discussion} 
\begin{figure}[tb]\vspace{-0.2cm}
\centering
\includegraphics[width=5.5cm]{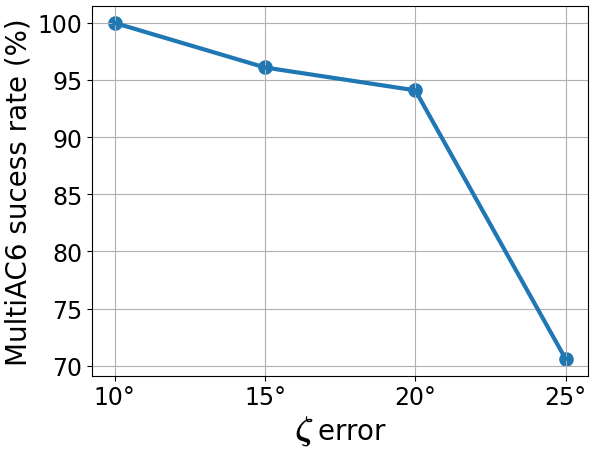}\vspace{-3pt}
\caption{\squeezeup \small \Update MultiAC6 success rate with respect to $\boldsymbol{\zeta}$ error. \Done}
\label{zeta_error}\squeezeup\squeezeup
\end{figure}

The simulation and experimental results clearly emphasize the benefits of actuating the 6 DOF of a gripper. These results suggest that controlling the gripper orientation is necessary, but not sufficient, to close the sim-to-real gap. By exploiting the gripper orientation within the MultiAC6 framework, the robot can achieve, with the same model, complex deformations for various DLOs. To do so, the desired orientation of the DLO tip $\boldsymbol{\zeta}$ is required to define the state of $\text{Agent}_o$. \Update This $\boldsymbol{\zeta}$ can be obtained empirically without the dynamic model of the DLO. \Done We acknowledge that this may be impractical for real-world deployment. This limitation could be related to many methods that require online fine-tuning \cite{Yu2023, Lagneau2020, Reza2022}. However, during our experiments, we noticed that MultiAC6 could accommodate coarse values of $\boldsymbol{\zeta}$ to achieve the desired DLO deformation. Taking advantage of the robustness of MultiAC6, the same orientation $\boldsymbol{\zeta}$ can be transferred to different DLOs without affecting real-world performance. \Update Furthermore, these orientations $\boldsymbol{\zeta}$ can be defined in advance without accurate measurements (see Figure~\ref{zeta_error}). Therefore, we believe that our approach may be less restrictive than online fine-tuning. 
Further limitations of MultiAC6 are related to discrete actuation peculiar to DRL. Discrete actuation can induce jerky motion and delays. Fortunately, these can be mitigated with longer time steps and interpolated velocities. \Done
\squeezeup

\vspace{-0.2cm} 
\section{Conclusion}

This article introduced MultiAC6, a new multi Actor-Critic framework to control large 3D deformations of DLOs with a single-arm robot. MultiAC6 decomposes the action space of a robot on different agents: one agent controls the gripper position and another controls the gripper orientation. The learning process is then simplified, since both the action and the state spaces are reduced. MultiAC6 was validated through extensive experiments in simulation and in the real world. The results proved that MultiAC6 can perform large deformations of up to 40 cm in a real setup. Furthermore, MultiAC6 is able to handle several types of DLO without retraining or online fine-tuning. We validated the robustness of MultiAC6 in real experiments using various unknown DLOs with an average success rate of 95\%.
\Update
In the future, we wish to develop new DRL frameworks to test MultiAC6 on other soft objects than DLOs and make new comparisons.
\Done


\section*{ACKNOWLEDGMENT}
This work is funded by the EU Horizon 2020 research and innovation programme under grant agreement No 101017284 (Project `ACROBA’) and by the French government through the France 2030 programme IdEx université de Bordeaux / RRI ROBSYS. \squeezeup \vspace{-0.2cm}

\bibliographystyle{ieeetr}
\bibliography{biblio}

\end{document}